# Sheep identity recognition, age and weight estimation datasets

Aya Salama Abdelhady[1], Aboul Ella Hassanenin[*], and Aly Fahmy[1]

[1]Faculty of Computer Science and Information, Cairo University, Cairo, Egypt
[*]Scientific Research Group in Egypt (SRGE) Cairo University, Cairo, Egypt
www.egyptscience.net

**Abstract:** Increased interest of scientists, producers and consumers in sheep identification has been stimulated by the dramatic increase in population and the urge to increase productivity. The world population is expected to exceed 9.6 million in 2050. For this reason, awareness is raised towards the necessity of effective livestock production. Sheep is considered as one of the main of food resources. Most of the research now is directed towards developing real time applications that facilitate sheep identification for breed management and gathering related information like weight and age. Weight and age are key matrices in assessing the effectiveness of production. For this reason, visual analysis proved recently its significant success over other approaches. Visual analysis techniques need enough images for testing and study completion. For this reason, collecting sheep images database is a vital step to fulfill such objective. We provide here datasets for testing and comparing such algorithms which are under development. Our collected dataset consists of 416 color images for different features of sheep in different postures. Images were collected fifty two sheep at a range of year from three months to six years. For each sheep, two images were captured for both sides of the body, two images for both sides of the face, one image from the top view, one image for the hip and one image for the teeth. The collected images cover different illumination, quality levels and angle of rotation. The allocated data set can be used to test sheep identification, weigh estimation, and age detection algorithms. Such algorithms are crucial for disease management, animal assessment and ownership.

**Keywords**: Sheep recognition, sheep weights estimation, Sheep age estimation, sheep data sets

## 1. Introduction

Sheep is important livestock species in developing countries. Sheep contribute to a broad range of production systems. Sheep farming has become one of the major commodities in the world. Variety of technologies is needed now to achieve better performance. Main goal of sheep farming is to make high profit by good management applications. Sheep identification is vital for breed management, disease management, assessment and ownership. On the other side, the size and age of sheep are related to their productivity [1]. Heavy weighted animals usually produce more meat and are more valuable. Weight and age are also the main keys for evaluation of the growth performance. Moreover, these features are the basic factor in deciding which animals to buy, sell, cull or mate. Manually techniques are used to estimate weight like scales. However, farms lack proper maintenance for this equipment. Using scales are also very hectic and time consuming to carry sheep to the scale frequently. Furthermore, consumers are usually pruned to manipulation. As a result, real time simple application is needed for weight, identity and age estimation. As a result, estimation of identity, weight and age based on visual analysis is considered as a hot topic now days. Visual analysis proved its reliability in many related application [2]. Age is estimated by the farmers based on their teeth shape. Therefore, they can be estimated by teeth image. Linear measurements of animals and body size can also be used to estimate weight [3]. Approaches based on visual analysis are reviewed in the following sections.

## 1.1 Sheep recognition using facial biometric

Facial images are the most common biometric characteristic used for recognition from still images or video [4]. Facial recognition depends on attributes such as eyes, eyebrows, nose, lips, and chin. Facial recognition has been used for sheep identification using independent-components algorithm [5]. Cosine distance classifier has been used. The recognition rate was 95.3% . Fewer independent components reduced recognition rate. However, higher number of training images improved accuracy to 96%. This experiment was done on fifty facial images

## 1.2 Weight estimation using visual analysis

Dual camera system was used to estimate weight and size of live sheep [6]. These two high resolution cameras were connected with a laptop and managed in Matlab 7.1. They experimented on inly 27 sheep of Alpagota type. In this study linear measurements were used for weigh estimation. Sheep height, body length and chest depth were measured to determine the size and weight of sheep. Partial Least Square was then used to get the relationship between these measurement and weight. The efficiency has been evaluated by the mean size error. For withers height and chest depth, error was around 3.5% and 5.0% for body length

## 2. Methods

Data collection of sheep images is the most important step for sheep recognition, weight and age estimation using visual analysis. The collected data/images were collected under different transformations: illumination, rotation, quality levels, and image partiality. Moreover, the images were taken from different instances and in different sizes. We were keen that side and top view images of sheep are having black background. These images are usually used in research for weight estimation. Black standard background helps researchers in segmentation and background subtraction.

Fifty two sheep faces, body and teeth were imaged over a period of five weeks. The sheep of different ages were captured, from 3 month to 6 years old. They were of Baraa breed which is Libyan breed but rose in a farm in Ismalia in Egypt. Sheep were selected in different physiological and growing. Pregnant, non pregnant, lactating, adults and still growing. Each sheep was restrained by a rope from their legs so as not to affect the images collected. Digital images of the frontal face were taken at a distance of average distance 1 m using high resolution mobile camera with a resolution of $1024 \times 768$ pixels. 8 photos were acquired per individual sheep. Two face images, two body images, and image from the top view, one image for the hip and one image for the teeth per sheep. To have a total of 416 color images. These images can be used for the training stage for facial recognition, weight and age estimation algorithms. The sheep acquisition system tem was built as shown in Figure 1. Images are taken as mentioned before under different transformations of illumination, rotation, quality levels, and image partiality. Table 1 lists the number of samples for each category.

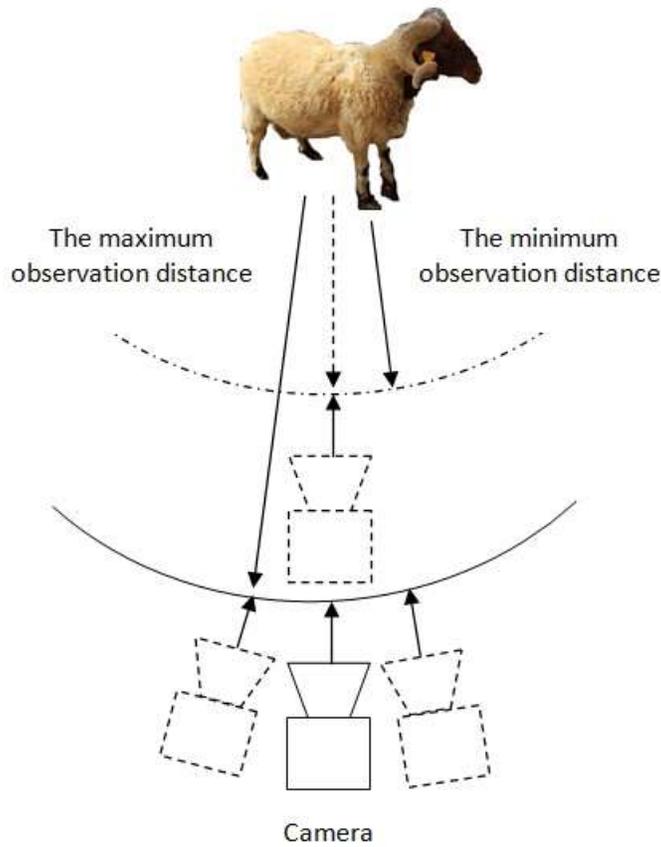
Figure 1: Sheep acquisition system

Table 1: Resources used for data collection

| | |
|---|---|
| Subject area | Computer Science |
| More specific subject area | Image processing, Machine Learning, and animal Biometrics. |
| Type of data | Image. |
| How the data was acquired | Professional cameraof Samsung smart |
| Data format | Jpeg image. |
| Experimental factors | No sample pre-processing applied. |
| Experimental features | 416 Image. |
| Data source location | Animal farm in Ismailia. |
| Data accessibility | Data are available with this article. |

Table 2: Usage of the collected data set

| Value of the data |
|---|
| The dataset is used to build computational model for sheep recognition. |
| The dataset is used to build computational model for sheep weight estimation. |
| The dataset is used to build computational model for sheep age estimation. |

## 3. Data Records

The data set consists of two images for each face side, two images for each body side, hip image and the teeth image as shown in figure 2. Data collected can be categorized into 5 categories as shown in the following table. Sheep were selected in different physiological and growing, Pregnant, non pregnant, lactating, adults, still growing, males and females. These images were collected under different transformations: illumination, rotation, quality levels, distances and image partiality.

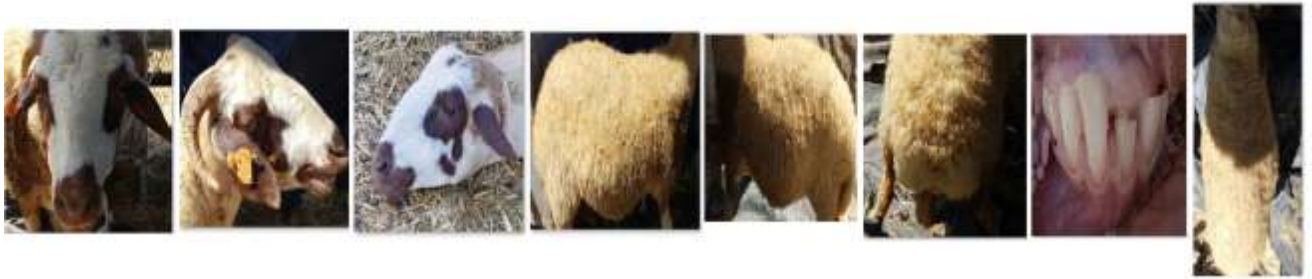

Figure 2: The different images taken for each Arabian horse

Table 3: Categories of the Collected Data

| Data category | Number of images |
|---|---|
| sheep | 52 |
| Sheep side image | 104 |
| Sheep top image | 70 |
| Hip image | 52 |
| Face | 104 |
| All images | **416** |

### (A) First category (Side, top and hip sheep images for weight estimation)

As we mentioned before, visual analysis has already proved in many related area for animal management. Side images and top view have already been used before for weigh estimation. For beef weight estimation, side picture of cattle were used. Then morphological operation and image segmentation were applied to extract the cattle. The number of pixels of this extracted contour was the feature key to determine the carcass weight of cattle [7]. Top images were also used in another research for pigs weight estimation. The top view areas were captured for the whole pig. Head, ears and neck were cropped as pre-processing. A regression equation was then used to predict weight from the relation with measured area. Girth dimension was also measured as another predictor of weight. Girth is shown in figure 3. For these reasons, we collected side images, top view images and hip images for the sheep to be used in the different approaches for weigh estimation. We collected the images at different illumination condition and slightly variant distance and rotation to cover all the possible conditions as shown in figure 4. As obvious in figure 4, we were keen that all of these images are having black background for easier segmentation and background subtraction.

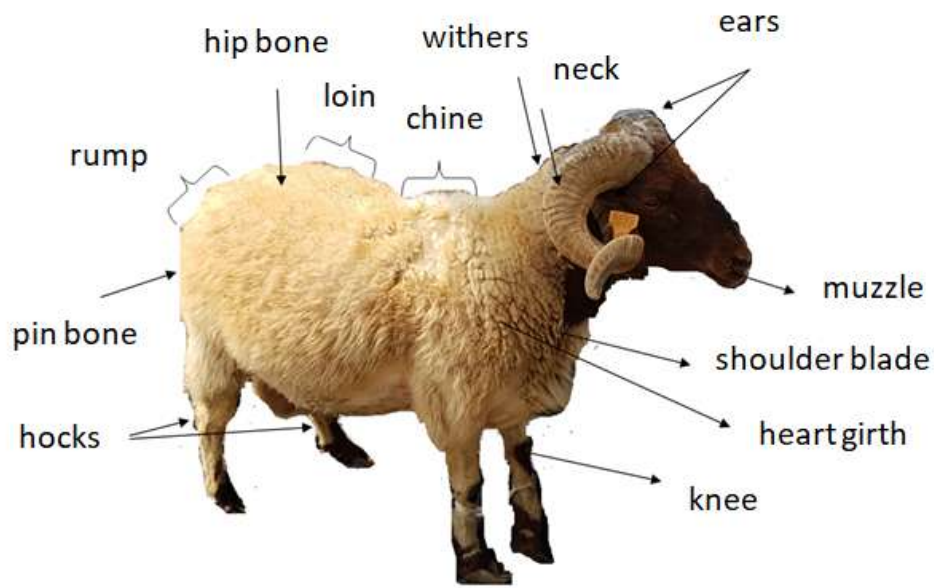

*Figure 3. Body parts of the sheep*

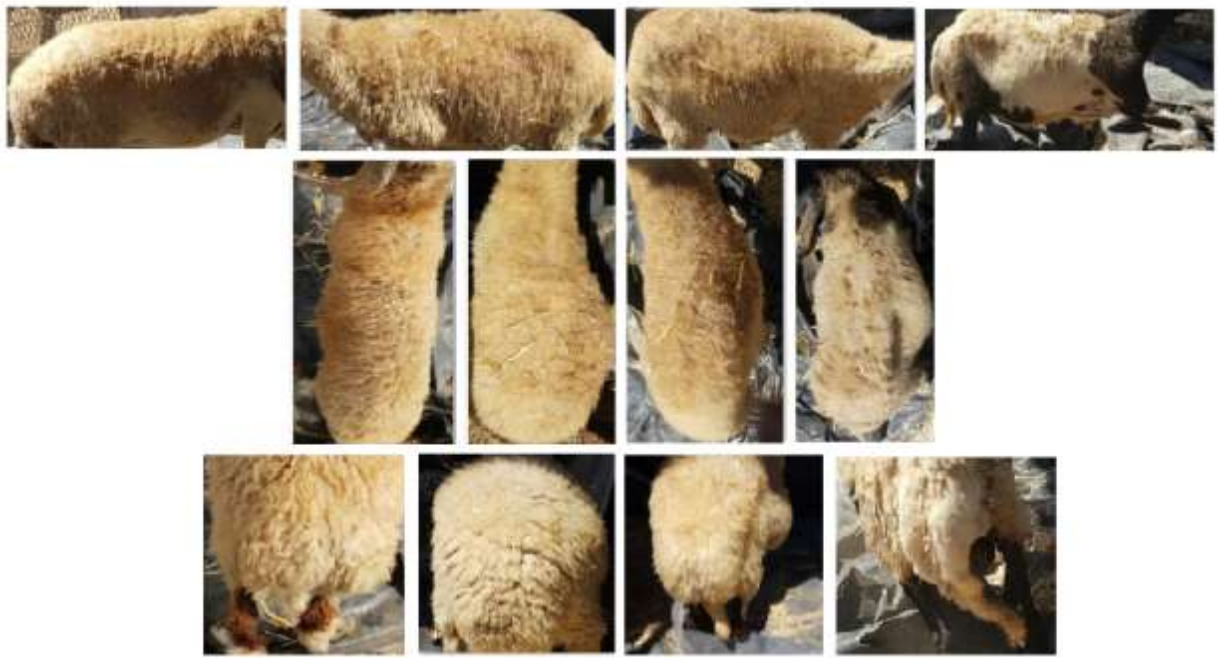

**Figure 4: Images taken for weight estimation**

Figure 4 shows side, top view and hip images taken from different angles. They were also taken at different illumination conditions.

**(B) Second category ( Face and side face images for sheep recognition)**

Face is one the main biometrics used for recognition as we mentioned before wither for human or for the animals. Muscle and eyes also can be used for the recognition algorithms and they also proved their feasibility to be used in related work [8]. For this reason, we captured different images for front face and two side images for the left and right as illustrated in figure 5. As a result, our data set can be used to train different algorithm for muscle biometric, eye biometric and whole face biometric.

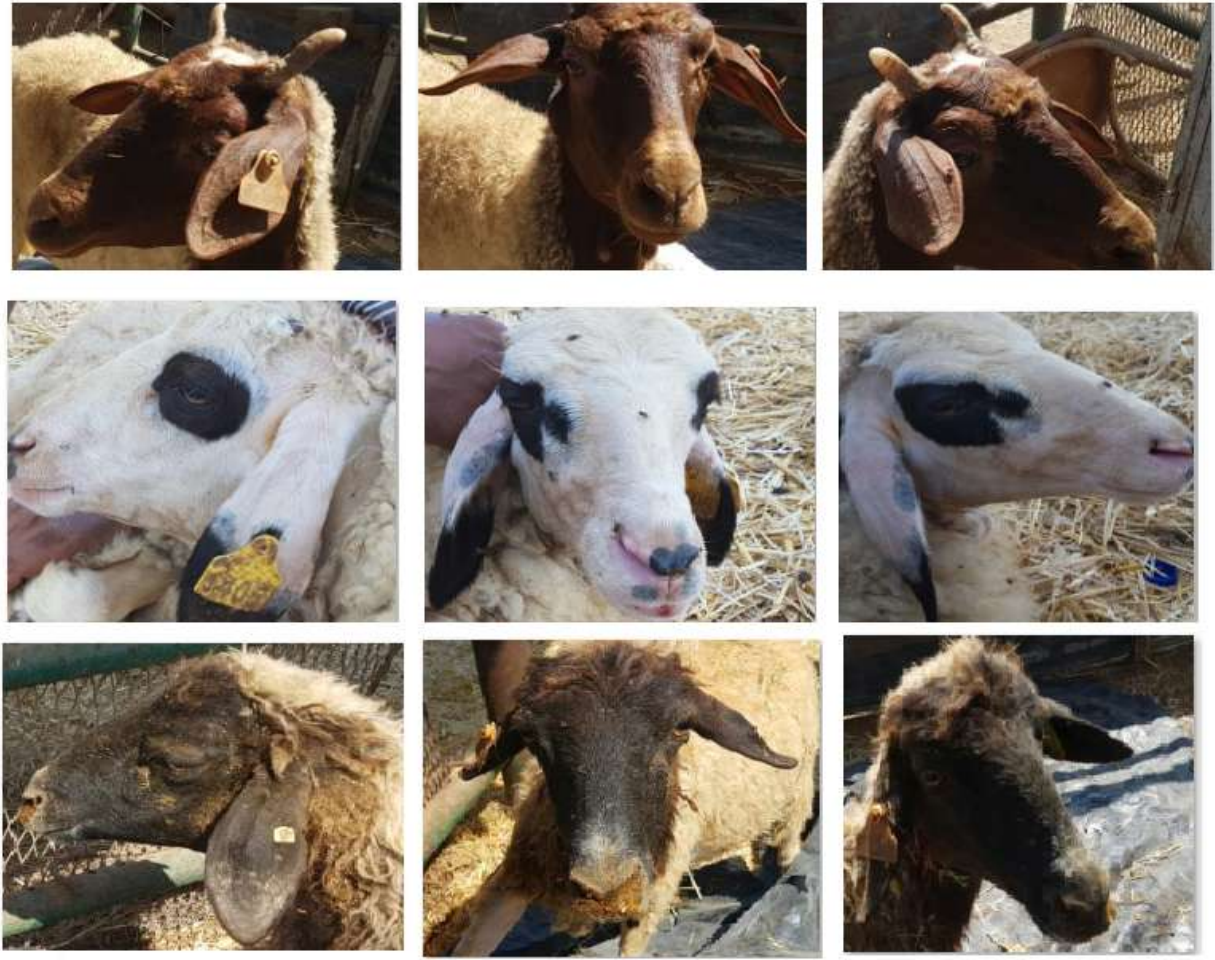

**Figure 5: Images taken for face recognition**

**(C) Third category ( Sheep teeth images for age detection)**

Farmers usually do not keep records animals age. As a result, farmers rely on dentition for age estimation. For this reason, we captured images of the teeth for each individual sheep as shown in Figure 6. We cleaned the teeth of some sheep for clear images and we left the dirt for other to cover different noisy conditions.

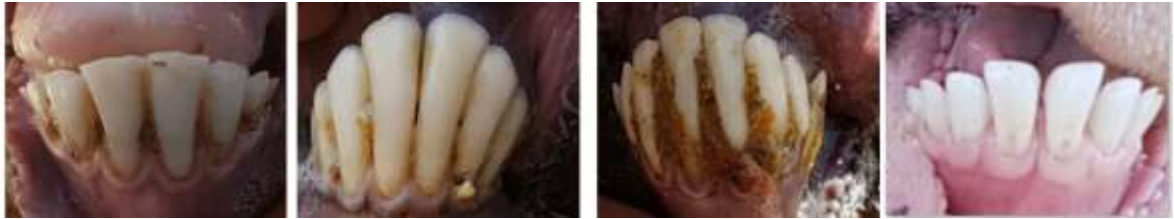

**Figure 6: A sample of teeth for age detection**

**Acknowledgement**

We would like to thank the staff working on the 6[th] October farm – ismalia, for helping to collect the images from the farm